\pdfoutput=1

\documentclass[11pt]{article}

\usepackage[]{acl}

\usepackage{times}
\usepackage{latexsym}

\usepackage[T1]{fontenc}

\usepackage[utf8]{inputenc}

\usepackage{microtype}

\usepackage{bm}
\usepackage{amsmath}
\usepackage{url}
\usepackage{graphicx}
\usepackage{float} 
\usepackage{subfigure}
\usepackage{multirow}
\usepackage{makecell}
\usepackage{xpinyin}
\usepackage{enumerate}

\usepackage{xcolor}
\usepackage{bbding}
\usepackage{soul}
\newcolumntype{L}[1]{>{\raggedright\let\newline\\\arraybackslash\hspace{0pt}}m{#1}}
\setul{2pt}{2pt}
\usepackage{booktabs}
\usepackage{amsmath,amsfonts}
\usepackage[utf8]{inputenc}
\usepackage{cleveref}

\crefname{section}{§}{§§}
\Crefname{section}{§}{§§}
\usepackage{enumitem}
\setenumerate[1]{itemsep=0pt,partopsep=0pt,parsep=\parskip,topsep=5pt}
\setitemize[1]{itemsep=0pt,partopsep=0pt,parsep=\parskip,topsep=5pt}
\setdescription{itemsep=0pt,partopsep=0pt,parsep=\parskip,topsep=5pt}
\usepackage{arydshln}

\usepackage[linesnumbered,ruled,vlined]{algorithm2e}

\title{MSCTD: A Multimodal Sentiment Chat Translation Dataset}

\author{
  Yunlong Liang\textsuperscript{1,2}\thanks{ \ \ Equal contribution. Work was done when Yunlong were interning at Pattern Recognition Center, WeChat AI, Tencent Inc, China.}  ,  
  Fandong Meng\textsuperscript{2}\footnotemark[1]  , 
  \textbf{Jinan Xu}\textsuperscript{1}\thanks{ \ \ Jinan Xu is the corresponding author.}  , 
  \textbf{Yufeng Chen}\textsuperscript{1}\ and \textbf{Jie Zhou}\textsuperscript{2}\\
  \textsuperscript{1}Beijing Key Lab of Traffic Data Analysis and Mining, \\Beijing Jiaotong University, Beijing, China \\
  \textsuperscript{2}Pattern Recognition Center, WeChat AI, Tencent Inc, China \\
  \texttt{\{yunlongliang,jaxu,chenyf\}@bjtu.edu.cn} \\
  \texttt{\{fandongmeng,withtomzhou\}@tencent.com} \\
}

\begin{document}
\maketitle
\begin{abstract}
Multimodal machine translation and textual chat translation have received considerable attention in recent years. Although the conversation in its natural form is usually multimodal, there still lacks work on multimodal machine translation in conversations. In this work, we introduce a new task named \textbf{M}ultimodal \textbf{C}hat \textbf{T}ranslation (MCT), aiming to generate more accurate translations with the help of the associated dialogue history and visual context. To this end, we firstly construct a \textbf{M}ultimodal \textbf{S}entiment \textbf{C}hat \textbf{T}ranslation \textbf{D}ataset (MSCTD) containing 142,871 English-Chinese utterance pairs in 14,762 bilingual dialogues and 30,370 English-German utterance pairs in 3,079 bilingual dialogues. Each utterance pair, corresponding to the visual context that reflects the current conversational scene, is annotated with a sentiment label. Then, we benchmark the task by establishing multiple baseline systems that incorporate multimodal and sentiment features for MCT. Preliminary experiments on four language directions (English$\leftrightarrow$Chinese and English$\leftrightarrow$German) verify the potential of contextual and multimodal information fusion and the positive impact of sentiment on the MCT task. Additionally, as a by-product of the MSCTD, it also provides two new benchmarks on multimodal dialogue sentiment analysis. Our work can facilitate research on both multimodal chat translation and multimodal dialogue sentiment analysis.\footnote{The code, data, and image features are publicly available at: \url{https://github.com/XL2248/MSCTD}}
\end{abstract}

\section{Introduction}
Multimodal machine translation~\cite{huang-etal-2016-attention,calixto-liu-2017-incorporating} and textual chat translation~\cite{wang-etal-2016-automatic,farajian-etal-2020-findings,liang-etal-2021-modeling} mainly focus on investigating the potential visual features and dialogue context, respectively. Both of them have received much attention. Although plenty of studies on them have been carried out based on either image captions~\cite{calixto-etal-2017-doubly,calixto-etal-2019-latent,ive-etal-2019-distilling,yin-etal-2020-novel,yao-wan-2020-multimodal} or textual dialogues~\cite{wang-etal-2017-semantics,maruf-etal-2018-contextual,liang-etal-2021-towards}, to our knowledge, little research work has been devoted to multimodal machine translation in conversations. One important reason is the lack of multimodal bilingual conversational datasets.
\textbf{\begin{figure*}[t]
    \centering
    \includegraphics[width=0.98\textwidth]{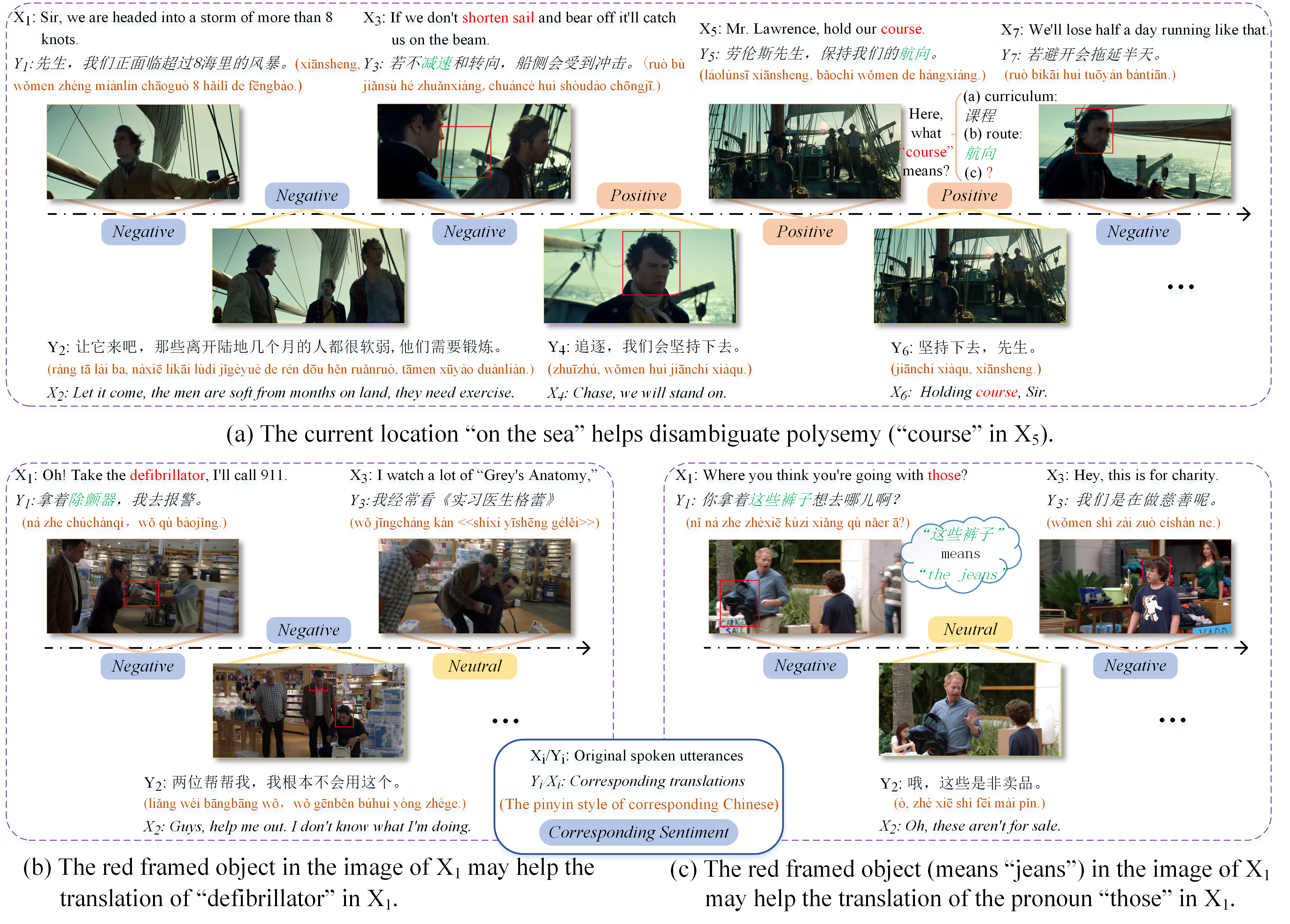}
    \caption{Three examples of the annotated multimodal bilingual dialogue in our MSCTD and the conversation is going from left to right.
    }
    \label{fig.1}
\end{figure*}}

Generally, conversation in its natural form is multimodal~\cite{poria-etal-2019-meld,liang2020infusing}. When humans converse, what a speaker would say next depends largely on what he/she sees. That is, the visual information plays a key role in (\emph{\romannumeral1}) supplementing some crucial scene information (\emph{e.g.}, the specific locations or objects, or facial expressions), (\emph{\romannumeral2}) resolving ambiguous multi-sense words (\emph{e.g.}, bank), and (\emph{\romannumeral3}) addressing pronominal anaphora issues (\emph{e.g.}, it/this). For instance, as shown in~\autoref{fig.1} (a), the image obviously points out the current location ``on the sea'', which may help disambiguate the meaning of ``course'' in the utterance $\text{X}_5$. Specifically, the dialogue history (\emph{i.e.}, talking about maritime affairs) and the corresponding visual context (\emph{i.e.}, on the sea/boat) assist us to determine that the word ``course'' means ``route/direction'' instead of ``curriculum''. In~\autoref{fig.1} (b), the visual context indicates object information, \emph{i.e.}, the ``defibrillator'' in $\text{X}_1$, which may help with translation. In~\autoref{fig.1} (c), the image of the utterance $\text{X}_1$ also demonstrates that it can provide appropriate candidates (\emph{i.e.,} the jeans) when translating the pronoun ``these''. Besides, the image offers some clues to judge the sentiment when it is hard to judge the polarity based only on the utterance (\emph{e.g.}, $\text{Y}_2$ in~\autoref{fig.1} (b) and $\text{X}_3$ in~\autoref{fig.1} (c)). All of the above call for a real-life multimodal bilingual conversational data resource that can encourage further research in chat translation.

In this work, we propose a new task named \textbf{M}ultimodal \textbf{C}hat \textbf{T}ranslation (MCT), with the goal to produce more accurate translations by taking the dialogue history and visual context into consideration. To this end, we firstly construct a \textbf{M}ultimodal \textbf{S}entiment \textbf{C}hat \textbf{T}ranslation \textbf{D}ataset (MSCTD). The MSCTD includes over 17k multimodal bilingual conversations (more than 142k English-Chinese and 30k English-German utterance pairs), where each utterance pair corresponds with the associated visual context indicating where it happens. In addition, each utterance is annotated with one sentiment label (\emph{i.e.}, positive/neutral/negative). 

Based on the constructed MSCTD, we benchmark the MCT task by establishing multiple Transformer-based~\cite{vaswani2017attention} systems adapted from several advanced representative multimodal machine translation models~\cite{ive-etal-2019-distilling,yao-wan-2020-multimodal} and textual chat translation models~\cite{ma-etal-2020-simple,liang-etal-2021-towards}. Specifically, we incorporate multimodal features and sentiment features into these models for a suitable translation under the current conversational scene. Extensive experiments on four language directions (English$\leftrightarrow$Chinese and English$\leftrightarrow$German) in terms of BLEU~\cite{papineni2002bleu}, METEOR~\cite{denkowski-lavie-2014-meteor} and TER~\cite{snover2006study}, demonstrate the effectiveness of contextual and multimodal information fusion, and the positive impact of sentiment on MCT. Furthermore, experiments on the multimodal dialogue sentiment analysis task of the three languages show the added value of the proposed MSCTD. 

In summary, our main contributions are:
\begin{itemize}

\item We propose a new task: multimodal chat translation named MCT, to advance multimodal chat translation research.
\item We are the first that contributes the human-annotated multimodal sentiment chat translation dataset (MSCTD), which contains 17,841 multimodal bilingual conversations, totally 173,240 <English utterance, Chinese/German utterance, image, sentiment> quadruplets. 

\item We implement multiple Transformer-based baselines and provide benchmarks for the new task. We also conduct comprehensive analysis and ablation study to offer more insights.

\item As a by-product of our MSCTD, it also facilitates the development of multimodal dialogue sentiment analysis.  
\end{itemize}
\section{Tasks}
\label{sec:td}
In this section, we firstly clarify the symbol definition, and then define the proposed~\emph{Multimodal Chat Translation} task and the existing~\emph{Multimodal Dialogue Sentiment Analysis} task. 

In a multimodal bilingual conversation (\emph{e.g.}, ~\autoref{fig.1} (a)), we assume the two speakers have alternatively given utterances in different languages for $u$ turns, resulting in $X_1, X_2, X_3, X_4,..., X_u$ and $Y_1, Y_2, Y_3, Y_4,...,Y_u$ on the source and target sides, respectively, along with the corresponding visual context representing where it happens: $Z_1, Z_2, Z_3, Z_4,..., Z_u$. Among these utterances, $X_1, X_3, X_5,..., X_u$ are originally spoken by the first speaker and $Y_1, Y_3, Y_5,..., Y_u$ are the corresponding translations in the target language. Similarly, $Y_2, Y_4, Y_6,..., Y_{u-1}$ are originally spoken by the second speaker and $X_2, X_4, X_6,..., X_{u-1}$ are the translated utterances in the source language. According to languages and modalities, we define three types of context: (1) the dialogue history context of $X_u$ on the source side as $\mathcal{C}_{X_u}$=$\{X_1, X_2, X_3,..., X_{u-1}\}$, and (2) that of $Y_u$ on the target side as $\mathcal{C}_{Y_u}$=$\{Y_1, Y_2, Y_3,..., Y_{u-1}\}$, and (3) the visual dialogue context $\mathcal{C}_{Z_u}$=$\{Z_1, Z_2, Z_3,..., Z_{u-1}, Z_{u}\}$.\footnote{For each item of \{$\mathcal{C}_{X_u}$, $\mathcal{C}_{Y_u}$\}, we add the special token `[CLS]' tag at the head of it and use another token `[SEP]' to delimit its included utterances, as in~\newcite{bert}.}

\paragraph{Multimodal Chat Translation.}
When translating the $u$-th utterance $X_u$=$\{x_{u,1}, x_{u,2}, ..., x_{u,N}\}$, the goal of the MCT task is to generate $Y_u$=$\{y_{u,1}, y_{u,2}, ..., y_{u,T}\}$ with the guidance of bilingual dialogue history contexts $\mathcal{C}_{X_u}$ and $\mathcal{C}_{Y_u}$ and the associated visual context $\mathcal{C}_{Z_u}$. Formally, the probability distribution of the target utterance $Y_u$ is defined as follows:
\begin{equation}
\label{eq:dnmt}
        {P}(Y_{u}|X_u, \mathcal{C}_{u}) = \prod_{t=1}^{T}p(y_{u,t}|y_{u,<t}, X_{u}, \mathcal{C}_{u}),
\end{equation}
where $y_{u,<t}$ = $\{y_{u,1}, y_{u,2}, y_{u,3}, ..., y_{u,t-1}\}$ and $\mathcal{C}_{u}$=$\{\mathcal{C}_{X_u}, \mathcal{C}_{Y_u}, \mathcal{C}_{Z_u}\}$.
\paragraph{Multimodal Dialogue Sentiment Analysis.}
Taking the $u$-th utterance $X_u$ for example, the task aims to predict a sentiment label $\ell \in $\{\emph{Positive}, \emph{Neutral}, \emph{Negative}\} for it given the corresponding image $Z_u$ and the dialogue history $\mathcal{C}_{X_u}$. 

\begin{table*}[htbp]
\centering
\scalebox{0.85}{
\setlength{\tabcolsep}{1.10mm}{
\begin{tabular}{c|l|rrrrccrrr}
\hline
\bf{MSCTD}&\bf{Type}  & \#\bf{Dial.} & \#\bf{Utter.} & \#\bf{Images} &\#\bf{AvgTurns} & \#\bf{AvgEn} & \#\bf{AvgZh/De} & \#\bf{Pos.} & \#\bf{Neu.} & \#\bf{Neg.} \\
\hline
\multirow{3}{*}{{Chinese$\rightarrow$English}} 
&Train    &13,749 &62,593 &62,593  &9.65&8.35 &10.84 &16,902 &24,074 &21,617\\
&Valid    &504 &2,389 &2,389 &10.05&8.27 &10.84 &708  &809 &872 \\
&Test     &509 &2,385 &2,385 &9.95&8.13 &11.09 &756  &618 &1,011 \\\cdashline{1-11}[4pt/2pt]
\multirow{3}{*}{{English$\rightarrow$Chinese}} 
&Train    &13,749 &70,148 &70,148 &9.65&8.34 &10.84 &18,478 &27,762 &23,908 \\
&Valid    &504 &2,674 &2,674  &10.05&8.14 &10.93 &746  &955 &973\\
&Test     &509 &2,682 &2,682 &9.95&8.19 &10.97 &850  &680 &1,152\\\cdashline{1-11}[4pt/2pt]
\multirow{3}{*}{{German$\rightarrow$English}} 
&Train    &2,066 &9,561 &9,561  &9.80&8.39 &8.46 &2,581 &3,281 &3,699\\
&Valid    &504 &2,389 &2,389 &10.05&8.27 &8.17 &708  &809 &872 \\
&Test     &509 &2,385 &2,385 &9.95&8.13 &8.36 &756  &618 &1,011 \\\cdashline{1-11}[4pt/2pt]
\multirow{3}{*}{{English$\rightarrow$German}} 
&Train    &2,066 &10,679 &10,679 &9.80&8.40 &8.45 &2,902 &3,640 &4,137 \\
&Valid    &504 &2,674 &2,674  &10.05&8.14 &8.14 &746  &955 &973\\
&Test     &509 &2,682 &2,682 &9.95&8.19 &8.29 &850  &680 &1,152\\\cdashline{1-11}[4pt/2pt]
&T./A.    &17,841 &173,241 &173,241 &9.91&8.25 &10.91/8.31 &46,983  &64,881 &61,376\\
\hline
\end{tabular}}}
\caption{Detailed Statistics of our MSCTD. \#: number of the corresponding item, \emph{i.e.}, Dial.: dialogues; Utter.: utterances; AvgTurns: Average turn length of each dialogue; AvgEn: Average length of each turn in English (word level); AvgZh/De: Average length of each turn in Chinese (character level) and in German (word level); Pos./Neu./Neg.: positive/neutral/negative sentiment label. The ``T./A.'' means the \underline{T}otal number or \underline{A}verage value of each column.}
\label{Tbl:data}
\end{table*}

\section{Dataset}
In this section, we mainly introduce our MSCTD in five aspects: \emph{Data Source}~\autoref{ds}, \emph{Annotation Procedure}~\autoref{ap}, \emph{Annotation Quality Assessment}~\autoref{aqa}, \emph{Dataset Statistics}~\autoref{dsa}, and the introduction of \emph{Related Datasets}~\autoref{rds}.
\subsection{Data Source}
\label{ds}
We mainly select the multimodal dialogues from the public available OpenViDial dataset~\cite{meng2021openvidial}, where each monolingual (English) utterance corresponds to an image. Since the original English utterance in OpenViDial is automatically extracted from the corresponding movie image by optical character recognition (OCR)\footnote{https://github.com/JaidedAI/EasyOCR}, it contains a lot of noises or errors. Furthermore, the lack of associated translations and sentiment labels for utterances, makes it impossible for directly conducting research on multimodal chat translation, sentiment-aware machine translation, and multimodal dialogue sentiment analysis with this data. Therefore, we further correct the wrong English utterances and annotate the corresponding Chinese/German translations and sentiment labels.

\subsection{Annotation Procedure}
\label{ap}

To build the MSCTD, the annotation procedure includes two steps: automatic annotation and then human annotation according to the annotation rules. 

\noindent\textbf{Automatic Annotation.} To improve the annotation efficiency, we firstly construct a paired English-Chinese subtitle database\footnote{To build this database, we firstly crawl two consecutive English and Chinese movie subtitles (not aligned) from here~\url{https://www.kexiaoguo.com/}. Then, we use several advanced technologies (\emph{e.g.}, Vecalign~\cite{thompson-koehn-2019-vecalign} and LASER~\cite{schwenk-2018-filtering}) to align these subtitles. Finally, we obtain the large-scale bilingual dialogue dataset (28M). We will also release this dataset, together with the MSCTD, to facilitate subsequent research.}. 
Then, we utilize the original English utterance to automatically select its Chinese translation by perfectly matching the English subtitle in the constructed bilingual database. As a result, about 78.57\% original English utterances are paired with Chinese translations.

\noindent\textbf{Human Annotation.} Since the full data are large, we divide the data into three parts and employ three annotators who are Chinese postgraduate students highly proficient in English comprehension. Each annotator is responsible for annotating one part according to the following guidelines: 
\begin{itemize}
\item Check and correct each English utterance;
\item Check and correct the matched Chinese subtitle to suit the current conversational scene;
\item For the remaining 21.43\% (without Chinese subtitles), translate them according to the corrected English utterance, the corresponding image, and the dialogue history.
\end{itemize}
Additionally, we employ another three annotators to label sentiment polarity for each utterance {independently} (\emph{i.e.}, each one annotates the full data) according to the current utterance, the associated image and the dialogue history. Following~\citet{firdaus-etal-2020-meisd}, majority voting scheme is used for selecting the final sentiment label for each utterance.

Finally, having the conversations in both languages allows us to simulate bilingual conversations where one speaker speaks in English and the other responds in Chinese~\cite{farajian-etal-2020-findings,liang-etal-2021-modeling}. ~\autoref{fig.1} shows three bilingual conversations where the two speakers have alternatively given utterances, along with their corresponding translations. By doing so, we build the MSCTD\footnote{For English$\leftrightarrow$German, we firstly sample a small set of training data and apply the same test and validation set with the English$\leftrightarrow$Chinese version. Then, the German translations are collected from professional English-German workers contracted via a language service company (\href{https://www.magicdatatech.com/}{magicdatatech}). The three crwodworkers are asked to translate them according to the English utterance, the corresponding image, and the dialogue history.}.  

\begin{table*}[ht!]
\centering
\scalebox{0.72}{
\setlength{\tabcolsep}{0.950mm}{
\begin{tabular}{l|cccc|rrr|rrrr}
\hline
\multirow{2}{*}{\textbf{Dataset}} &\multirow{2}{*}{\textbf{Language Direction}} &\multirow{2}{*}{\textbf{Modality}}&\multirow{2}{*}{\textbf{Scene}} &\multirow{2}{*}{\textbf{Sentiment}} &  \multicolumn{3}{c|}{\textbf{\#Dialogues}}  &  \multicolumn{3}{c}{\textbf{\#Instances/Utterances}} \\
\cline{6-8} \cline{9-11}
 & & & & &Train & Valid & Test &Train & Valid & Test\\
\hline
{Multi30K}~\cite{elliott-etal-2016-multi30k}   & English$\rightarrow$German/French  & T,V   &Caption &\XSolidBrush  &-     &- &-  &29,000 &1,014    & 1,000  \\\cdashline{1-11}[4pt/2pt]
{BSD-AMI-ON}~\cite{rikters-etal-2020-document}  & English$\leftrightarrow$Japanese &T &Dialogue  &\XSolidBrush  &2,643   &69 &69    &84,800  &2,058    &2,104      \\{BconTrasT}~\cite{farajian-etal-2020-findings}  & English$\leftrightarrow$German &T  &Dialogue &\XSolidBrush  &550   &78 &78    &13,845  &1,902    &2,100      \\
{BMELD}~\cite{liang-etal-2021-modeling}   &English$\leftrightarrow$Chinese   &T &Dialogue &\XSolidBrush  &1,036 &108 &274     &9,987 &1,084 &2,601      \\\cdashline{1-11}[4pt/2pt]
\multirow{2}{*}{MSCTD (Ours)}  &English$\leftrightarrow$Chinese   &T,V & Dialogue  &\Checkmark     &13,749 & 504     &509 &132,741  &5,063   &5,067       \\
&English$\leftrightarrow$German   &T,V & Dialogue  &\Checkmark     &2,066 & 504     &509 &20,240  &5,063   &5,067       \\
\hline
\end{tabular}}}
\caption{Comparison of (1) previous mulitmodal machine translation dataset: Multi30k, (2) textual chat translation datasets: BconTrasT, BSD-AMI-ON, and BMELD, and (3) our MSCTD. T/V: text/vision modality.  } 
\label{data:tr}
\end{table*}

\subsection{Annotation Quality Assessment}
\label{aqa}
To evaluate the quality of annotation, we use Fleiss' Kappa to measure the overall annotation consistency among three annotators~\cite{doi:10.1177/001316447303300309}. We measure this data from two aspects: translation quality and sentiment quality. 

For translation quality, we measure the inter-annotator agreement on a subset of data (sample 50 dialogues with 504 utterances), and we ask the three annotators mentioned above to re-annotate this subset {independently}. Then, we invite another postgraduate student to measure the inter-annotator agreement on the re-annotated subset by the three annotators. Finally, the inter-annotator agreement calculated by Fleiss’ kappa are 0.921 for English$\leftrightarrow$Chinese and 0.957 for English$\leftrightarrow$German, respectively. They indicate ``Almost Perfect Agreement'' between three annotators.

For sentiment quality, we measure the inter-annotator agreement on the full dataset. The inter-annotator agreements calculated by Fleiss’ kappa is 0.695, which indicates ``Substantial Agreement'' between three annotators. The level is consistent with previous work~\cite{firdaus-etal-2020-meisd} which can be considered as reliable.

\subsection{Dataset Statistics}
\label{dsa}
As shown in~\autoref{Tbl:data}, the MSCTD contains totally 17,841 bilingual conversations and 142,871/30,370 English-Chinese/English-German utterance pairs with two modalities (\emph{i.e.}, text and image), where each utterance has been annotated with onesentiment label. For English-Chinese/English-German, we split the dialogues into 13,749/2,066 for train, 504/504 for valid, and 509/509 for test while keeping roughly the same distribution of the utterance pair/image, respectively. The detailed annotation of sentiment labels are also listed in~\autoref{Tbl:data}, where three labels account for similar proportion. 

Based on the statistics in~\autoref{Tbl:data}, the average number of turns per dialogue is about 10, and the average numbers of tokens per turn are 8.2, 10.9, and 8.3 for English utterances (word level), Chinese utterances (character level), and German utterance (word level), respectively. 

\subsection{Related Datasets}
\label{rds}
The related datasets mainly involve three research fields: multimodal machine translation, textual chat translation, and multimodal dialogue sentiment analysis. 

In \textbf{multimodal machine translation}, there exists one dataset: Multi30K~\cite{elliott-etal-2016-multi30k}, where each image is paired with one English caption and two human translations into German and French. It is an extension of the original English description dataset: Flickr30K~\cite{young-etal-2014-image}. Afterwards, some small-scale multimodal test sets (about 3k instances) are released to evaluate the system, such as WMT18 test set (1,071 instances)~\cite{barrault-etal-2018-findings}.

In \textbf{textual chat translation}, three datasets have been released: BSD-AMI-ON~\cite{rikters-etal-2020-document}, BconTrasT~\cite{farajian-etal-2020-findings}, and BMELD~\cite{liang-etal-2021-modeling}. The BSD-AMI-ON is a document-aligned Japanese-English conversation corpus, which contains three sub-corpora: Business Scene Dialogue (BSD~\cite{rikters-etal-2019-designing}), Japanese translation of AMI meeting corpus (AMI~\cite{mccowan2005ami}), and Japanese translation of OntoNotes 5.0 (ON~\cite{Weischedel2017OntoNotesA}). The BconTrast and BMELD are two human-annotated datasets, which are extended from monolingual textual dialogue datasets Taskmaster-1~\cite{byrne-etal-2019-taskmaster} and MELD~\cite{poria-etal-2019-meld}, respectively.

In \textbf{multimodal dialogue sentiment analysis}, the MELD~\cite{poria-etal-2019-meld} and MEISD~\cite{firdaus-etal-2020-meisd} datasets are publicly available. The MELD dataset is constructed by extending the EmotionLines~\cite{hsu-etal-2018-emotionlines} from the scripts of the popular sitcom \emph{Friends}. It is similar to MEISD, which is also built from famous English TV shows under different genres (\emph{e.g.}, \emph{Friends}, \emph{Grey's Anatomy}, \emph{The Big Bang Theory}).

The resources mentioned above are extensively used in corresponding fields of research and they even cover some sub-tasks in MSCTD. However, our MSCTD is different from them in terms of both complexity and quantity. 

Firstly, multimodal machine translation datasets and textual chat translation datasets are either in multimodal or textual dialogue, while ours includes both. It is obvious that conducting multimodal machine translation in conversations is more challenging due to the more complex scene. Furthermore, MSCTD covers four language directions and contains more than 17k human-annotated utterances-image triplets, which is more than the sum of the annotated ones in Multi30K, BSD-AMI-ON, BconTrasT, and BMELD. ~\autoref{data:tr} provides information on the number of available modality, dialogues, and their constituent utterances for all the five datasets. What is more, our MSCTD is also annotated with sentiment labels while they are not. 

Secondly, compared with two existing multimodal dialogue sentiment analysis datasets, MSCTD's quantity of English version is nearly ten-times of the annotated utterances in MEISD or MELD. More importantly, our MSCTD provides an equivalent Chinese multimodal dialogue sentiment analysis dataset and a relatively small German counterpart. ~\autoref{data:sa} shows the comparison for all the five datasets, \emph{i.e.}, MELD, MEISD, and our MSCTD on three languages. 

\begin{table}[t]
\centering
\newcommand{\tabincell}[2]{\begin{tabular}{@{}#1@{}}#2\end{tabular}}
\small
\scalebox{0.8}{
\setlength{\tabcolsep}{0.8mm}{
\begin{tabular}{l|rrr|rrr}
\hline
\multirow{2}{*}{\bf{Dataset}} & \multicolumn{3}{c|}{\#\bf{Dialogues}} &  \multicolumn{3}{c}{\#\bf{Utterances}} \\
\cline{2-4} \cline{5-7}
&Train &Valid & Test&Train &Valid & Test\\\hline

MELD~\cite{poria-etal-2019-meld}   &1,039&114&280 &9,989 &1,109 &2,610\\
MEISD~\cite{firdaus-etal-2020-meisd}   &702&93&205   &14,040&1,860&4,100 \\\cdashline{1-7}[4pt/2pt]
{MSCTD-Zh} (Chinese version)   &13,749 & 504     &509 &132,741  &5,063   &5,067\\
{MSCTD-En} (English version)   &13,749 & 504     &509 &132,741  &5,063   &5,067\\
{MSCTD-De} (German version)    &2,066 & 504     &509  &20,240  &5,063   &5,067\\
\hline
\end{tabular}}}
\caption{Comparisons of four multimodal dialogue sentiment analysis datasets: MELD, MEISD, and our MSCTD on two languages.}\label{data:sa}
\end{table}

\section{Image Features}
Following previous work~\cite{wang-etal-2018-object,ive-etal-2019-distilling,meng2021openvidial}, we focus on two types of image representation, namely the coarse-grained spatial visual feature maps and the fine-grained object-based visual features.

\paragraph{Coarse-grained Spatial Visual (CSV) Features.} We use the ResNet-50 model~\cite{7780459} pre-trained on ImageNet~\cite{5206848} to extract a high-dimensional feature vector $\mathbf{f}_j \in \mathbb{R}^{d_c}$ for image $Z_j$. These features contain output activations for various filters while preserving spatial information. We refer to models that use such features as \textbf{CSV}.

\paragraph{Fine-grained Object-based Visual (FOV) Features.} Since using coarse-grained image features may be insufficient to model fine-grained visual elements in images including the specific locations, objects, and facial expressions, we use a bag-of-objects representation where the objects are obtained using an off-shelf Faster R-CNNs~\cite{NIPS2015_14bfa6bb} pre-trained on Visual Genome~\cite{krishnavisualgenome}. Specifically, for an input image $Z_j$, we obtain a set of detected objects from Faster R-CNNs, \emph{i.e.}, $\mathbf{O}_j$ = $\{\mathbf{o}_{j,1}, \mathbf{o}_{j,2}, \mathbf{o}_{j,3}, ..., \mathbf{o}_{j,m}\}$, where $m$ is the number of extracted objects and $\mathbf{o}_{j,*} \in \mathbb{R}^{d_f}$. Each object is captured by a dense feature representation, which can be mapped back to a bounding box / region (\emph{i.e.}, Region-of-Interest (ROI)). We refer to models that use such features as \textbf{FOV}.

Both types of features have been used in various vision and language tasks such as multimodal dialogue sentiment analysis~\cite{firdaus-etal-2020-meisd}, image captioning~\cite{pmlr-v37-xuc15,shi-etal-2021-enhancing}, and multimodal machine translation~\cite{ive-etal-2019-distilling,lin2020dynamic,SU202147}.

\begin{table*}[ht!]
\centering
\scalebox{0.70}{
\setlength{\tabcolsep}{0.48mm}{
\begin{tabular}{c|l|l|lcl|lcl|lcl|lcl}
\hline
\multirow{2}{*}{\textbf{Modality}}&\multirow{2}{*}{M\#} &\multirow{2}{*}{\textbf{Model}}&  \multicolumn{3}{c|}{\textbf{Chinese$\rightarrow$English}}  &  \multicolumn{3}{c|}{\textbf{English$\rightarrow$Chinese}} &  \multicolumn{3}{c|}{\textbf{German$\rightarrow$English}} &  \multicolumn{3}{c}{\textbf{English$\rightarrow$German}} \\
\cline{4-6} \cline{7-9} \cline{10-12} \cline{13-15}
&&  & BLEU$\uparrow$ & METEOR$\uparrow$ & TER$\downarrow$  &BLEU$\uparrow$ & METEOR$\uparrow$ & TER$\downarrow$& BLEU$\uparrow$ & METEOR$\uparrow$ & TER$\downarrow$  &BLEU$\uparrow$ & METEOR$\uparrow$ & TER$\downarrow$\\
\hline
\multirow{3}{*}{\textbf{T}} 
&M1\ \;\quad &{Trans.}    &19.98    &23.46\;\,  &61.55    & 24.66  &25.25\;\;\,     &60.39   &21.74 &27.87\;\,&57.27&21.46 &22.91\;\,&60.50\\
&M2\ \;\quad&{TCT}              &20.39    &24.28\;\,  &61.32    & 25.21  &25.79\;\;\,     &60.17    &21.99&27.98\;\,&57.71&21.77&23.22\;\,&60.35 \\ 
&M3\ \;\quad&{CA-TCT}              &20.83    &24.67\;\,  &60.84    & 25.62  &26.05\;\;\,     &59.37  &22.27&28.03\;\,&56.82&22.19&23.71\;\,&59.58\\\cdashline{1-15}[4pt/2pt]
\multirow{5}{*}{\textbf{T+CSV}}
&M4\ \;\quad&{Trans.+Emb}$^*$     &21.03     &24.44\;\,  &60.54     &25.51 &26.04\;\;\,     &59.79     &21.94&27.94\;\,&56.70&22.54&23.04\;\,&58.96\\
&M5\ \;\quad&{Trans.+Sum}$^*$     &21.29     &25.06\;\,  &60.43     &26.06 &26.33\;\;\,     &58.57      &21.99&27.98\;\,&56.56&22.02&23.08\;\,&59.51\\
&M6\ \;\quad&{Trans.+Att}$^*$     &21.54     &25.24\;\,  &60.35     &26.10 &26.48\;\;\,     &58.29      &23.00&28.53\;\,&56.52&22.72&23.51\;\,&58.04 \\\textbf{}
&M7\ \;\quad&{MCT} (Ours)       &22.00     &25.46\;\,  &59.85     &26.54  &26.75\;\;\,   &58.07    &23.34&28.71\;\,&56.33&23.12&23.94\;\,&58.57 \\ 
&M8\ \;\quad&{CA-MCT} (Ours)                                           &\underline{22.51}$^{\dagger\ddagger}$     &25.50$^{\dagger}$ &59.34$^{\dagger\ddagger}$    &\underline{26.83}$^{\dagger\ddagger}$  &\underline{26.97}$^{\dagger\ddagger}$   &57.72$^{\dagger\ddagger}$ &23.81$^{\dagger\ddagger}$&28.94$^{\ddagger}$&55.67$^{\dagger\ddagger}$&23.48$^{\ddagger}$&\underline{24.21}$^{\ddagger}$&58.33\\\cdashline{1-15}[4pt/2pt]
\multirow{5}{*}{\textbf{T+FOV}}
&M9\ \;\quad&{Trans.+Con}$^*$      &21.53     &24.87\;\,   &59.56     &25.47 &26.18\;\;\,     &59.00   &22.17&28.26\;\,&56.02&22.19&23.25\;\,&58.07    \\
&M10\quad&{Trans.+Obj}$^*$     &21.82     &25.35\;\,   &59.99     &26.24 &26.42\;\;\,    &57.92     &22.41&28.73\;\,&55.42&22.88&23.64\;\,&\underline{57.46} \\
&M11\quad&{M-Trans.}$^*$        &22.38     &25.77\;\,   &\underline{59.15}   &26.60  &26.65\;\;\,     &57.84  &23.40&29.10\;\,&55.21&23.18&24.00\;\,&57.71 \\
&M12\quad&{MCT} (Ours)  &22.46     &\underline{25.88}\;\,  &59.27     &26.74 &26.83\;\;\,      &\underline{57.59}   &\underline{23.94}&\underline{29.19}\;\,&\underline{55.03}&\underline{23.79}&24.16\;\,&57.65  \\ 
&M13\quad&{CA-MCT} (Ours) &\textbf{22.87}$^{\dagger\ddagger}$    &\textbf{25.94}$^{\dagger}$ &\textbf{58.57}$^{\dagger\ddagger}$     &\textbf{27.04}$^{\dagger\ddagger}$   &\textbf{27.12}$^{\dagger\ddagger}$   &\textbf{57.56}$^{\dagger\ddagger}$ &\textbf{24.33}$^{\dagger\ddagger}$&\textbf{29.42}$^{\dagger}$&\textbf{54.90}$^{\dagger}$&\textbf{24.12}$^{\dagger\ddagger}$&\textbf{24.41}$^{\dagger}$&\textbf{57.24}$^{\dagger}$\\
\hline
\end{tabular}}}
\caption{Test results of multimodal chat translation task in terms of BLEU, METEOR, and TER on our MSCTD. The best and the second results are \textbf{bold} and \underline{underlined}, respectively. The symbol `$^*$' denotes sentence-level multimodal machine translation models which do not use the dialogue history. `$^{\dagger}$' indicates that statistically significant better than the M3 model with t-test {\em p} \textless \ 0.01.  `$^{\ddagger}$' indicates that statistically significant better than the sentence-level multimodal machine translation models (\emph{i.e.}, M13 vs. M11 and M8 vs. M6) with t-test {\em p} \textless \ 0.05.} 
\label{tbl:main_res}
\end{table*}
\section{Baseline Models}

To provide convincing benchmarks for the MSCTD, we perform experiments with multiple Transformer-based~\cite{vaswani2017attention} models for the multimodal chat translation task. Additionally, we provide several baselines for the multimodal dialogue sentiment analysis task. 
\subsection{Multimodal Chat Translation}
According to different visual features, we divide the baselines into three categories: text only (\textbf{T}), text plus coarse visual features (\textbf{T + CSV}), and text plus fine-grained visual features (\textbf{T + FOV}).

\textbf{T}: Trans.~\cite{vaswani2017attention}: the standard transformer model, which is a sentence-level neural machine translation (NMT) model~\cite{yanetal2020multi,meng2019dtmt,zhangetal2019bridging}, \emph{i.e.}, regardless of the dialogue history. {TCT}~\cite{ma-etal-2020-simple}: A unified document-level NMT model based on Transformer by sharing the first encoder layer to incorporate the dialogue history, which is used as the Textual Chat Translation (TCT) model by~\cite{liang-etal-2021-towards}.
{CA-TCT}~\cite{liang-etal-2021-towards}: A multi-task learning model that uses several auxiliary tasks to help model generate coherence-aware translations.

\textbf{T+CSV}:
Trans.+Emb~\cite{vaswani2017attention}: it concatenates the image feature to the word embedding and then trains the sentence-level NMT model. Trans.+Sum~\cite{ive-etal-2019-distilling}: it adds the projected image feature to each position of the encoder output. Trans.+Att~\cite{ive-etal-2019-distilling}: this model utilizes an additional cross-attention sub-layer to attend the image features in each decoder block. MCT: we implement the multimodal self-attention~\cite{yao-wan-2020-multimodal} in the encoder to incorporate the image features into the chat translation model. CA-MCT: similarly, we incorporate image features into the multitask-based chat translation model~\cite{liang-etal-2021-towards} by the multimodal self-attention. 

\textbf{T+FOV}:
Trans.+Con~\cite{vaswani2017attention}: it concatenates the word sequence to the extracted object sequence and thus obtains a new sequence taken as the input of the sentence-level NMT model. Trans.+Obj~\cite{ive-etal-2019-distilling}: it is a translate-and-refine model (two-stage decoder) where the images are only used by a second-pass decoder. M-Trans.~\cite{yao-wan-2020-multimodal}: it leverages a multimodal self-attention layer to encode multimodal information where the hidden representation of images are induced from the text under the guidance of image-aware attention. MCT: here, we incorporate the object-level features into the model instead of coarse one. CA-MCT: similarly, we incorporate the object-level features into the multi-task learning model.

\subsection{Multimodal Dialogue Sentiment Analysis}
We perform several experiments with different models. 
text-CNN~\cite{kim-2014-convolutional}: it only applies CNNs to extract textual information for each utterance in a dialogue. In this approach, we do not use the dialogue history or the additional visual information. DialogueRNN~\cite{majumder2019dialoguernn}: this baseline is a powerful approach for capturing dialogue history with effective mechanisms for sentiment analysis.
DialogueRNN + BERT~\cite{firdaus-etal-2020-meisd}: this model improves the performance of DialogueRNN by using BERT~\cite{bert} embeddings instead of Glove~\cite{pennington-etal-2014-glove} embeddings to represent the textual features. DialogueRNN + PLM: we propose a stronger baseline built upon the DialogueRNN for sentiment analysis. Specifically, we utilize RoBERTa~\cite{liu2019roberta} embeddings for English sentiment analysis, and ERNIE~\cite{sun2019ernie} embeddings for Chinese sentiment analysis, and XLM-R~\cite{conneau-etal-2020-unsupervised} embeddings for German sentiment analysis. 

Following~\citet{firdaus-etal-2020-meisd}, we only use the coarse-grained image features (\emph{i.e.}, \textbf{CSV}) when training above models with the visual information.

\section{Experiments}

\subsection{Setup}
\label{sect:data}
For {multimodal chat translation}, we utilize the standard Transformer-Base architecture~\cite{vaswani2017attention}. Generally, we use the settings described in previous work~\cite{ive-etal-2019-distilling,yao-wan-2020-multimodal,liang-etal-2021-towards} to conduct experiments on our MSCTD. 

For {multimodal dialogue sentiment analysis}, we mainly follow the settings of previous work~\cite{poria-etal-2019-meld,firdaus-etal-2020-meisd}. 

Please refer to~\autoref{ID} for more details.

\subsection{Metrics}
For {multimodal chat translation}, following previous work~\cite{liang-etal-2021-towards,ive-etal-2019-distilling}, we use the SacreBLEU\footnote{BLEU+case.mixed+numrefs.1+smooth.exp+tok.13a+\\version.1.4.13}~\cite{post-2018-call}, METEOR~\cite{denkowski-lavie-2014-meteor} and TER~\cite{snover2006study} with the statistical significance test~\cite{koehn-2004-statistical} for fair comparison. Specifically, for Chinese$\rightarrow$English, we report case-insensitive score. For English$\rightarrow$Chinese, the reported score is at the character level. For English$\leftrightarrow$German, we report case-sensitive BLEU score.

For {multimodal dialogue sentiment analysis}, following~\citet{poria-etal-2019-meld}, we report weighted-average F-score.

\subsection{Results of Multimodal Chat Translation}
\label{ssec:layout}

\paragraph{Results on English$\leftrightarrow$Chinese.}

(1) Among all only text-based models (M1$\sim$M3), we find that M1 performs worse than M2, showing that the dialogue history indeed is beneficial for better translations. Furthermore, M3 can further improve the translation performance, which suggests that modeling the coherence characteristic in conversations is crucial for higher results. These can also be found in other settings (\emph{e.g.}, M7 vs. M4$\sim$6; M8 vs. M7). (2) The models with image features incorporated get higher results than corresponding text-based models (\emph{i.e.}, M4$\sim$M6\&M9 vs. M1; M7\&M12 vs. M2; M8\&M13 vs. M3). (3) The dialogue history and the image features obtain significant cumulative benefits (M8 vs. M1 and M13 vs. M1) (4) Among these image-based models (M4$\sim$M8 or M9$\sim$M13), we observe that different fusion manners of text and image features reflect great difference on effects. It shows that there is much room for further improvement using other more advanced fusion methods. (5) Using \textbf{FOV} image features is generally better than the coarse counterpart \textbf{CSV} (M9$\sim$M13 vs. M4$\sim$M8), which demonstrates that the fine-grained object elements may offer more specific and effective information for better translations. 
\paragraph{Results on English$\leftrightarrow$German.}Similar findings are found on English$\leftrightarrow$German. This shows that our conclusions are solid and convincing on general datasets. All these results prove the value of our constructed MSCTD.

Furthermore, we provide some stronger baselines that we firstly train the model on the general-domain corpus and then fine-tune it on our chat translation dataset. The results are presented in Table~\autoref{tbl:p_main_res} of~\autoref{pftr}, which show similar findings observed in Table~\autoref{tbl:main_res}.

\begin{table}[t]
\centering
\newcommand{\tabincell}[2]{\begin{tabular}{@{}#1@{}}#2\end{tabular}}
\setlength{\tabcolsep}{0.8mm}{
\begin{tabular}{l|cccc}
\hline
\multirow{2}{*}{\textbf{Model}} & \multicolumn{3}{c}{$\textbf{Chinese$\rightarrow$English}$} \\
\cline{2-4} 
 &BLEU$\uparrow$ & METEOR$\uparrow$ & TER$\downarrow$\\
\hline
Transformer (\textbf{T})      &20.43 &24.06  &61.00 \\
TCT (\textbf{T})              &20.81 &24.45  &61.19 \\
CA-TCT (\textbf{T})           &21.23 &24.82  &60.75 \\\cdashline{1-4}[4pt/2pt]
MCT (\textbf{T+CSV})          &22.25   &25.60  &59.69 \\
CA-MCT (\textbf{T+CSV})       &22.68   &25.60  &59.14 \\
\hline
\end{tabular}}
\caption{Sentiment-aware translation results using ground truth.}
\label{sa_nmt} 
\end{table}

\begin{table}[t]
\centering
\newcommand{\tabincell}[2]{\begin{tabular}{@{}#1@{}}#2\end{tabular}}
\scalebox{0.85}{
\setlength{\tabcolsep}{0.5mm}{
\begin{tabular}{l|cccc}
\hline
\multirow{2}{*}{\textbf{Model}} & \multicolumn{4}{c}{$\textbf{Chinese$\rightarrow$English}$} \\
\cline{2-5} 
 &BLEU$\uparrow$ & METEOR$\uparrow$ & TER$\downarrow$&ACC.$\uparrow$\\
\hline
Transformer (\textbf{T})      &20.34 &24.01  &61.09 &64.17\\
TCT (\textbf{T})              &20.78 &24.39  &60.87 &64.65\\
CA-TCT (\textbf{T})           &21.15 &24.73  &60.74 &64.78\\\cdashline{1-5}[4pt/2pt]
MCT (\textbf{T+CSV})          &22.31   &25.42  &59.88&65.26 \\
CA-MCT (\textbf{T+CSV})       &22.57   &25.51  &59.45 &65.33\\
\hline
\end{tabular}}}
\caption{Sentiment-aware translation results using predicted sentiment labels. The last column (\emph{i.e.}, ACC.) is the corresponding predicted sentiment accuracy.}
\label{sa_nmt_p} 
\end{table}
\subsection{Effect of Sentiment on Multimodal Chat Translation}
\label{SMCT}
To evaluate the effect of sentiment, we conduct some experiments on several baselines including single-modality ones and double-modality ones. In terms of implementation, following~\citet{si-etal-2019-sentiment}, we append the sentiment label to the head of the source utterance. ~\autoref{sa_nmt} shows the results. Comparing them with the results (M1$\sim$M3 and M7$\sim$M8) without using the sentiment in~\autoref{tbl:main_res}, we find that using the ground-truth sentiment label has a positive impact on the translation performance. Therefore, we believe that it is a topic worthy of research in the future. 

We also conducted the experiments with automatically predicted sentiment labels rather than the gold ones as the reviewer suggested, where we used the mixed sentiment presentation by dot-multiplying the predicted sentiment distribution and the sentiment label representation. The results are shown in~\autoref{sa_nmt_p}, where we find that the sentiment factor, as the inherent property of conversations, indeed has a positive impact on translation performance. We also observe that using the automatically predicted sentiment labels (actually the mixed sentiment representation) shows slightly lower results than using ground truth in terms of three metrics. The reason may be that the mixed sentiment representation has certain fault tolerance.
\begin{table*}[ht!]
\centering
\scalebox{0.8}{
\setlength{\tabcolsep}{0.90mm}{
\begin{tabular}{l|cc|cccc|cccc|cccc}
\hline
\multirow{2}{*}{\textbf{Model}} &\multicolumn{2}{c|}{\textbf{Modality}} &  \multicolumn{4}{c|}{\textbf{MSCTD-Zh (Chinese)}}  &  \multicolumn{4}{c|}{\textbf{MSCTD-En (English)}} &  \multicolumn{4}{c}{\textbf{MSCTD-De (German)}} \\
\cline{2-3} \cline{4-7} \cline{8-11} \cline{12-15}
 & T & V & Pos. & Neu. & Neg. & W-avg.  & Pos. & Neu. & Neg. & W-avg.  & Pos. & Neu. & Neg. & W-avg.\\
\hline
text-CNN     &\checkmark &-     &52.69 &66.80 &60.49  & 61.19  &52.35  &63.33   &56.12   & 58.23 &42.49  &43.06   &55.55   & 48.21\\\cdashline{0-14}[4pt/2pt]
\multirow{3}{*}{DialogueRNN}  
&\checkmark &-          &52.16   &69.01    &58.77     &61.45     &52.40   &68.10  &55.91      &60.21 &36.38&39.52&62.88&48.50\\ 
&-   &\checkmark        &20.30   &37.61    &29.11      &28.49      &20.30   &37.61  &29.11    &28.49  &20.30   &37.61  &29.11    &28.49\\ 
&\checkmark&\checkmark  &51.83   &70.21    &60.76     &62.59        &55.51 &69.34  &57.81   &61.69 &41.88&46.67&55.36&49.15\\ \cdashline{0-14}[4pt/2pt]
\multirow{2}{*}{DialogueRNN+BERT}  
&\checkmark &-         &56.07  &73.64    &63.39     &65.57    &58.15  &71.90    &60.44    &64.40 &43.33&44.59&61.97&52.76\\
&\checkmark&\checkmark &57.38  &73.73    &65.73      &66.12    &59.15  &72.79    &61.63    &64.99 &43.14&46.40&61.87&53.32\\ \cdashline{0-14}[4pt/2pt]
\multirow{2}{*}{DialogueRNN+PLM$^*$ (Ours)}    
&\checkmark&-         &58.85 &73.86    &67.92      &67.18     &59.55  &73.30     &61.20    &65.94 &43.49&48.17&62.14&54.19  \\
&\checkmark&\checkmark&60.21 &74.13    &66.59     &\textbf{67.57}    &58.94  &74.27     &62.82    &\textbf{66.45} &44.36&48.14&62.51&\textbf{54.46}\\ 
\hline
\end{tabular}}}
\caption{Test results of multimodal dialogue sentiment analysis task in terms of weighted F-score (\%). The ``W-avg.'' denotes weighted-average F-score and the best ``W-avg.'' results are \textbf{bold}. The symbol `$^*$' denotes that we use pre-trained language models ERNIE, RoBERTa and XLM-R for Chinese, English, and German, respectively.  } 
\label{tbl:main_sa_res}
\end{table*}

\label{ssec:cs}
\subsection{Results of Multimodal Dialogue Sentiment Analysis}
In \autoref{tbl:main_sa_res}, we report the results of sentiment classification on three datasets under different settings. 

\paragraph{Results on MSCTD-Zh.}
\label{ssec:ende}
We can see that the text-based models perform much poorer than other multimodal systems, which shows that it is not enough to evaluate the sentiment based only on the text. It indicates that visual information and contextual embeddings are crucial for classifying sentiment polarities. Overall, we achieve weighted F1 score of 67.57\% with the ``DialogueRNN+ERNIE'' model. 

\paragraph{Results on MSCTD-En/MSCTD-De.}
\label{ssec:chen}
On English/German, we observe the same findings on Chinese. These show that it is beneficial to introduce the visual information and contextual embeddings into the multimodal dialogue sentiment analysis task for different languages. Overall, we achieve the best F1 score of 66.45\% and 54.46\% on English and German, respectively. 

On this task, we obtain consistent results with previous work~\cite{poria-etal-2019-meld,firdaus-etal-2020-meisd}, which suggests the utility and reliability of our MSCTD. Additionally, MSCTD-Zh and MSCTD-De bridge the gap on multimodal dialogue sentiment analysis of Chinese and German.

\section{Conclusion and Future Work}
In this paper, we introduce a new multimodal machine translation task in conversations. Then, we construct a multimodal sentiment chat translation dataset named MSCTD. Finally, we establish multiple baseline systems and demonstrate the importance of dialogue history and multimodal information for MCT task. Additionally, we conduct multimodal dialogue sentiment analysis task on three languages of the MSCTD to show its added value.

MCT is a challenging task due to the complex scene in the MSCTD, leaving much room for further improvements. This work mainly focuses on introducing the new task and dataset, and we provide multiple models to benchmark the task. In the future, the following issues may be worth exploring to promote the performance of MCT:
\begin{itemize}
\item How to effectively perceive and understand the visual scenes to better assist multimodal machine translation in conversations?
\item How to build a multimodal conversation representation model to effectively align, interact, and fuse the information of two modalities?
\end{itemize}

\section{Ethical Considerations}
\label{ec}
In this section, we discuss the main ethical considerations of MSCTD: (1) Intellectual property protection. The English utterance and image of MSCTD is from OpenViDial dataset~\cite{meng2021openvidial}. For our translation and sentiments, its permissions are granted to copy, distribute and modify the contents under the terms of the \href{https://en.wikipedia.org/wiki/Wikipedia:Text_of_Creative_Commons_Attribution-ShareAlike_3.0_Unported_License}{Creative Commons AttributionShareAlike 3.0 Unported License} and \href{https://www.wikidata.org/wiki/Wikidata:Text_of_the_Creative_Commons_Public_Domain_Dedication}{Creative Commons CC0 License}, respectively. (2) Privacy. The data source are publicly available movies. Its collection and Chinese/German annotation procedure is designed for chat translation purpose, and does not involve privacy issues. (3) Compensation. During the sentiment annotation, Chinese and German translation, the salary for annotating each utterance is determined by the average time of annotation and local labor compensation standard. (4) Data characteristics. We refer readers to the content and \newcite{meng2021openvidial} for more detailed characteristics. (5) Potential problems. While principled measures are taken to ensure the quality of the dataset, there might still be potential problems with the dataset quality, which may lead to incorrect translations in applications. However, moderate noise is common in large-scale modern translators, even for human translated sentences, which should not cause serious issues.

\section*{Acknowledgements}
The research work descried in this paper has been supported by the National Key R\&D Program of China (2020AAA0108001) and the National Nature Science Foundation of China (No. 61976015, 61976016, 61876198 and  61370130). The authors would like to thank the anonymous reviewers for their valuable comments and suggestions to improve this paper.

\bibliography{anthology,custom}
\bibliographystyle{acl_natbib}

\appendix
\label{sec:appendix}
\begin{table*}[ht!]
\centering
\scalebox{0.70}{
\setlength{\tabcolsep}{0.48mm}{
\begin{tabular}{c|l|l|lcl|lcl|lcl|lcl}
\hline
\multirow{2}{*}{\textbf{Modality}}&\multirow{2}{*}{M\#} &\multirow{2}{*}{\textbf{Model}}&  \multicolumn{3}{c|}{\textbf{Chinese$\rightarrow$English}}  &  \multicolumn{3}{c|}{\textbf{English$\rightarrow$Chinese}} &  \multicolumn{3}{c|}{\textbf{German$\rightarrow$English}} &  \multicolumn{3}{c}{\textbf{English$\rightarrow$German}} \\
\cline{4-6} \cline{7-9} \cline{10-12} \cline{13-15}
&&  & BLEU$\uparrow$ & METEOR$\uparrow$ & TER$\downarrow$  &BLEU$\uparrow$ & METEOR$\uparrow$ & TER$\downarrow$& BLEU$\uparrow$ & METEOR$\uparrow$ & TER$\downarrow$  &BLEU$\uparrow$ & METEOR$\uparrow$ & TER$\downarrow$\\
\hline
\multirow{4}{*}{\textbf{T}} 
&M1\ \;\quad &{Trans. w/o FT}    &19.18    &24.81\;\;\,  &60.72    & 25.79  &27.04\;\;\,     &61.24   &46.80 &38.99\;\;\,&35.19 &44.74 &35.42\;\,&35.48\\
&M2\ \;\quad &{Trans.}    &27.92    &29.31\;\;\,  &53.92    & 33.30  &29.79\;\;\,     &50.92   &50.57 &44.68\;\;\,&30.21&51.76 &39.11\;\,&29.96\\
&M3\ \;\quad&{TCT}              &28.28    &29.96\;\;\,  &51.42    & 33.94  &30.57\;\;\,     &50.85    &50.80&45.10\;\;\,&30.02&51.96&39.29\;\,&29.75 \\ 
&M4\ \;\quad&{CA-TCT}              &\underline{28.56}    &\underline{30.24}\;\;\,  &51.29    & \underline{34.42}  &\underline{31.11}\;\;\,     &\underline{50.21}  &51.27&45.21\;\;\,&29.88&52.24&39.36\;\,&\underline{29.52}\\\cdashline{1-15}[4pt/2pt]
\multirow{3}{*}{\textbf{T+FOV}}
&M5\ \;\quad&{Trans.+Con}$^*$      &28.16     &29.88\;\;\,   &51.65     &33.43 &30.10\;\;\,     &50.71   &51.52&45.87\;\;\,&29.33&52.18&39.45\;\,&29.67    \\
&M6\quad&{MCT} (Ours)  &28.49     &{30.11}\;\;\,  &\underline{51.28}     &34.07 &30.74\;\;\,      &{50.63}   &\underline{51.75}&\underline{45.91}\;\;\,&\underline{29.19}&\underline{52.45}&\underline{39.66}\;\,&29.55  \\ 
&M7\quad&{CA-MCT} (Ours) &\textbf{28.81}$^{\dagger\ddagger}$    &\textbf{30.45}$^{\dagger\ddagger}$ &\textbf{51.06}$^{\dagger\ddagger}$     &\textbf{34.77}$^{\dagger\ddagger}$   &\textbf{31.40}$^{\dagger\ddagger}$   &\textbf{50.05}$^{\dagger\ddagger}$ &\textbf{51.98}$^{\dagger\ddagger}$&\textbf{46.37}$^{\dagger\ddagger}$&\textbf{29.02}$^{\dagger\ddagger}$&\textbf{52.72}$^{\dagger\ddagger}$&\textbf{39.58}$^{\dagger}$&\textbf{29.39}$^{\dagger}$\\
\hline
\end{tabular}}}
\caption{Pre-training-then-fine-tuning results of multimodal chat translation task in terms of BLEU, METEOR, and TER on Test set. The best and the second results are \textbf{bold} and \underline{underlined}, respectively. The symbol `$^*$' denotes sentence-level multimodal machine translation models which do not use the dialogue history. `$^{\dagger}$' and `$^{\ddagger}$' indicates that statistically significant better than the M2 model with t-test {\em p} \textless \ 0.01 and {\em p} \textless \ 0.05, respectively.} 
\label{tbl:p_main_res}
\end{table*}
\section{Implementation Details}
\label{ID}
For {multimodal chat translation}, we utilize the standard Transformer-Base architecture~\cite{vaswani2017attention}. Generally, we use the settings described in previous work~\cite{ive-etal-2019-distilling,yao-wan-2020-multimodal,liang-etal-2021-towards} to conduct experiments on our MSCTD. Specifically, we use 512 as hidden size, 2048 as filter size and 8 heads in multihead attention. Both the encoder and the decoder of all the models have 6 layers and are trained using THUMT~\cite{tan-etal-2020-thumt}. We set the training step to 100,000 steps. The dropout is set to 0.1. The batch size for each GPU is set to 4096 tokens. The experiments are conducted using 4 NVIDIA Tesla V100 GPUs, which gives us about 4*4096 tokens per update. The models are optimized using Adam~\cite{kingma2017adam} with $\beta_1$=0.9 and $\beta_2$=0.998, and learning rate is set to 1.0. Label smoothing is set to 0.1. Following~\citet{liang-etal-2021-towards}, we set the number of dialogue context to 3. During inference, the beam size is set to 4, and the length penalty is 0.6 in all experiments.

For the pre-training-then-fine-tuning setting, we firstly train our model on the WMT20 datasets for 100,000 steps. Then, we utilize the pre-trained model to initialize our all multimodal chat translation models.

For {multimodal dialogue sentiment analysis}, we mainly follow the settings of previous work~\cite{poria-etal-2019-meld,firdaus-etal-2020-meisd,liang-etal-2021-iterative-multi,liang2020dependency}. The experiments are conducted on an NVIDIA Tesla V100 GPU and the batch size is set to 64. The learning rate is set to 0.001.

\section{Pre-training-then-fine-tuning Results}
\label{pftr}
In this section, we provide some stronger baselines that we firstly train the standard transformer~\cite{vaswani2017attention} model on the general-domain corpus (WMT20 dataset of~\autoref{wmt}) and then fine-tune it on our chat translation dataset (\emph{i.e.}, using the pre-training-then-fine-tuning paradigm). In~\autoref{tbl:p_main_res}, the M1 denotes we directly evaluate the pre-trained model on the target chat test set (\emph{i.e.}, without fine-tuning on chat translation dataset.). The M2$\sim$M7 apply the  pre-training-then-fine-tuning paradigm. From~\autoref{tbl:p_main_res}, we observe similar conclusions to~\autoref{ssec:layout}. This shows that our findings on the newly proposed dataset are solid even under the stronger baselines. Besides, we also find that, after pre-training on the general-domain corpus, the model obtains significant improvement (M2$\sim$M7 vs. M1).

\section{WMT20 Dataset} 
\label{wmt}
For English$\leftrightarrow$Chinese, we combine News Commentary v15, Wiki Titles v2, UN Parallel Corpus V1.0, CCMT Corpus, and WikiMatrix. For English$\leftrightarrow$German, we combine six corpora including Euporal, ParaCrawl, CommonCrawl, TildeRapid, NewsCommentary, and WikiMatrix. First, we filter out duplicate sentence pairs and remove those whose length exceeds 80. To pre-process the raw data, we employ a series of open-source/in-house scripts, including full-/half-width conversion, unicode conversation, punctuation normalization, and tokenization~\cite{wang-EtAl:2020:WMT1}. After filtering, we apply BPE~\cite{sennrich-etal-2016-neural} with 32K merge operations to obtain subwords. Finally, we obtain 22,244,006 sentence pairs for English$\leftrightarrow$Chinese and 45,541,367 sentence pairs for English$\leftrightarrow$German, respectively. 

\end{document}